\documentclass[journal]{IEEEtran}

\usepackage[ruled, linesnumbered, vlined, commentsnumbered]{algorithm2e}
\usepackage{amsfonts,amssymb,amsmath,amsthm,amsopn,mathrsfs,mathtools,wasysym}	
\usepackage{hhline,booktabs,colortbl,multirow,tabularx,diagbox,threeparttable} 
\usepackage[caption=false,font=normalsize,labelfont=sf,textfont=sf]{subfig}
\usepackage[listings,skins,breakable]{tcolorbox}
\usepackage{color,xcolor} 
\usepackage{graphicx} 
\usepackage{arydshln}
\usepackage{enumerate}
\usepackage{authblk}
\usepackage{footnote}
\usepackage{hyperref}
\usepackage{prettyref}
\usepackage{pifont}
\usepackage{cite}
\usepackage{bm}
\usepackage{pgf}
\usepackage{tikz}
\usepackage[activate={true,nocompatibility},final,tracking=true,kerning=true,spacing=true,factor=1100,stretch=10,shrink=10]{microtype}
\usepackage{setspace}
\usepackage{multirow}
\usepackage{fontawesome5}

\def\our{\texttt{LitLA}}

\usepackage[framemethod=TikZ]{mdframed}

\definecolor{mycyan}{gray}{.7}

\newtheorem{definition}{Definition}

\newenvironment{remark}[1]{
    \mdfsetup{%
        frametitle={%
            \tikz[baseline=(current bounding box.east),outer sep=0pt]
            \node[anchor=east,rectangle,fill=white,text=black]
            {\strut \textbf{\normalsize #1}};}, 
        innertopmargin=-3pt,
        linecolor=black,
        linewidth=1pt,
        topline=true,
        frametitleaboveskip=\dimexpr-\ht\strutbox\relax,
        font=\small, 
        skipabove=3pt, 
        skipbelow=3pt, 
    }
    \begin{mdframed}[]
}{
    \end{mdframed}
}

\newcommand{\pref}{\prettyref}

\newrefformat{fig}{Fig.~\ref{#1}}
\newrefformat{tab}{Table~\ref{#1}}
\newrefformat{sec}{Section~\ref{#1}}
\newrefformat{app}{Appendix~\ref{#1}}
\newrefformat{alg}{Algorithm~\ref{#1}}
\newrefformat{property}{Property~\ref{#1}}
\newrefformat{theorem}{Theorem~\ref{#1}}
\newrefformat{lemma}{Lemma~\ref{#1}}
\newrefformat{corollary}{Corollary~\ref{#1}}
\newrefformat{proposition}{Proposition~\ref{#1}}
\newrefformat{def}{Definition~\ref{#1}}
\newrefformat{eq}{equation~(\ref{#1})}
\definecolor{SoftSkyBlue}{RGB}{220,228,247}

\definecolor{NavyBlue}{RGB}{8,29,92}

\definecolor{set_blue}{RGB}{116,159,201}
\definecolor{set_light_blue}{RGB}{215,229,242}

\definecolor{set_light_orange}{RGB}{247,222,190}
\definecolor{set_mid_orange}{RGB}{244,197,154}
\definecolor{set_orange}{RGB}{243,174,123}

\definecolor{set_green}{RGB}{128,184,134}
\definecolor{set_green_light}{RGB}{194,226,184}
\definecolor{set_green_mid}{RGB}{219,238,211}

\definecolor{set_grey}{RGB}{138,138,138}
\definecolor{set_mid_grey}{RGB}{206,206,206}
\definecolor{set_light_grey}{RGB}{227,227,227}

\definecolor{set_purple}{RGB}{151,146,191}
\definecolor{set_mid_purple}{RGB}{183,180,211}
\definecolor{set_light_purple}{RGB}{228,228,239}

\newcommand{\coloredsquare}[1]{
\tikz \fill [#1] (0.5ex,0.5ex) rectangle (1.75ex,1.75ex);
}

\begin{document}

\title{Mapping Literature Landscapes with \\ Data-Driven Discovery: A Case Study on MOEA/D}

\author{Mingyu Huang, Shasha Zhou and
       Ke Li,~\IEEEmembership{Senior Member,~IEEE}
\thanks{M. Huang is with the School of Computer Science and Engineering, the University of Electronic Science and Technology of China, Chengdu, 611731, China.} 
\thanks{S. Zhou and K. Li are with the Department of Computer Science, University of Exeter, North Park Road, Exeter, EX4 4QF, UK.}
}

\markboth{} {Shell\MakeLowercase{\textit{et al.}}: Bare Demo of IEEEtran.cls for Journals}

\maketitle

\begin{abstract}
    We are living in an era of “big literature,” where scientific literature is expanding exponentially. While this growth presents new opportunities, it complicates mapping global scientific research landscapes, as manual review methods become infeasible. Recent advancements in machine learning, complex networks, and natural language processing have enabled numerous data-driven discovery methods. Building upon these tools, we introduce an end-to-end workflow for analyzing large-scale literature landscapes, \texttt{LitLA}. This workflow first integrates diverse publication metadata into a bibliographic knowledge graph (KG) representing the research landscape. It then offers tools for exploratory analysis of various landscape aspects. We demonstrate the effectiveness of \texttt{LitLA} via a case study on follow-up works of multi-objective evolutionary algorithm based on decomposition (MOEA/D). In doing so, we constructed the MOEA/D research landscape as a KG comprising over 5,400 papers, 10,000 authors, 1,600 institutions, and 78,000 keywords. With this landscape, we start with descriptive statistics and investigate prominent topics pertaining to MOEA/D and interrogate their spatial-temporal and bilateral relationships. We then map the collaboration and citation networks to reveal the community's growth over time. We further experiment whether learning on latent patterns of this landscape can hint on future research directions. 
\end{abstract}


%
\IEEEpeerreviewmaketitle

\section{Introduction}
\label{sec:introduction}

\IEEEPARstart{I}{n} the past decade, scientific literature has expanded at an unprecedented rate across many research areas. For example, publications on \lq genetic algorithm\rq\ or \lq evolutionary computation\rq\ since 2010 have increased nearly fourfold, reaching about $270,000$ records on \href{https://www.webofscience.com}{Web of Science} (WoS). By contrast, only $65,000$ related publications appeared in the first $35$ years following Holland’s original work on genetic algorithms in 1975~\cite{Holland75}. This surge in digital publication data offers exciting opportunities for mapping research trends. Yet, it also poses significant challenges for researchers and practitioners who strive to understand the evolution and structure of these fields. Typically, researchers conduct systematic literature reviews (SLRs)~\cite{KitchenhamBBTBL09} to summarize a topic by manually evaluating available studies. However, the exponential growth of publications makes it increasingly difficult to adhere to \textit{comprehensive}, \textit{objective}, \textit{open}, and \textit{transparent} (COOT) principles using traditional review methods. In many fields, standard surveys cover only a few hundred papers, which are just a fraction of the total. Thus, this inevitably introduces selection biases due to incomplete coverage.

At the same time, an influx of exploratory analytical methods for diverse data types has greatly assisted researchers in many disciplines. Techniques such as topic modeling~\cite{ChurchillS22} from natural language processing (NLP) help extract latent themes from large textual datasets, such as social media posts on climate change~\cite{FalkenbergGTDLMSPZQB22} and hotel reviews~\cite{AmeurHY24}. Community detection algorithms~\cite{FortunatoH16} from network science have uncovered social groups~\cite{BediS16} and functional brain regions~\cite{BrittinCHEC21}. Large language models (LLMs)~\cite{MinaeeMNCSAG} now automate labor-intensive tasks such as data annotation~\cite{TanBWGBJKLCL24}, data wrangling~\cite{HuhSC23}, and text summarization~\cite{ChangLGI24}. Meanwhile, scholarly databases like WoS, \href{https://www.scopus.com}{Scopus}, and \href{https://www.semanticscholar.org/}{Semantic Scholar} provide extensive bibliographic entries enriched with metadata such as abstracts, authorship data, publication venues, and citation information. This convergence of abundant bibliographic data and sophisticated data mining techniques offers opportunities for mapping entire research domains~\cite{LiuLLL13,HeldensHWMBN20,SunY17,JeongZD23}, as well as uncovering broader patterns in scientific structure and evolution~\cite{FortunatoBBEHMMRSUVWWB18,RossGMBEL22,SinatraWDSB16,WuWE19,LiACL19}. However, these approaches often appear in isolation when applied to literature exploration, and no comprehensive workflow has yet been proposed to integrate them.

\begin{figure*}[t!]
    \centering
    \includegraphics[width=\linewidth]{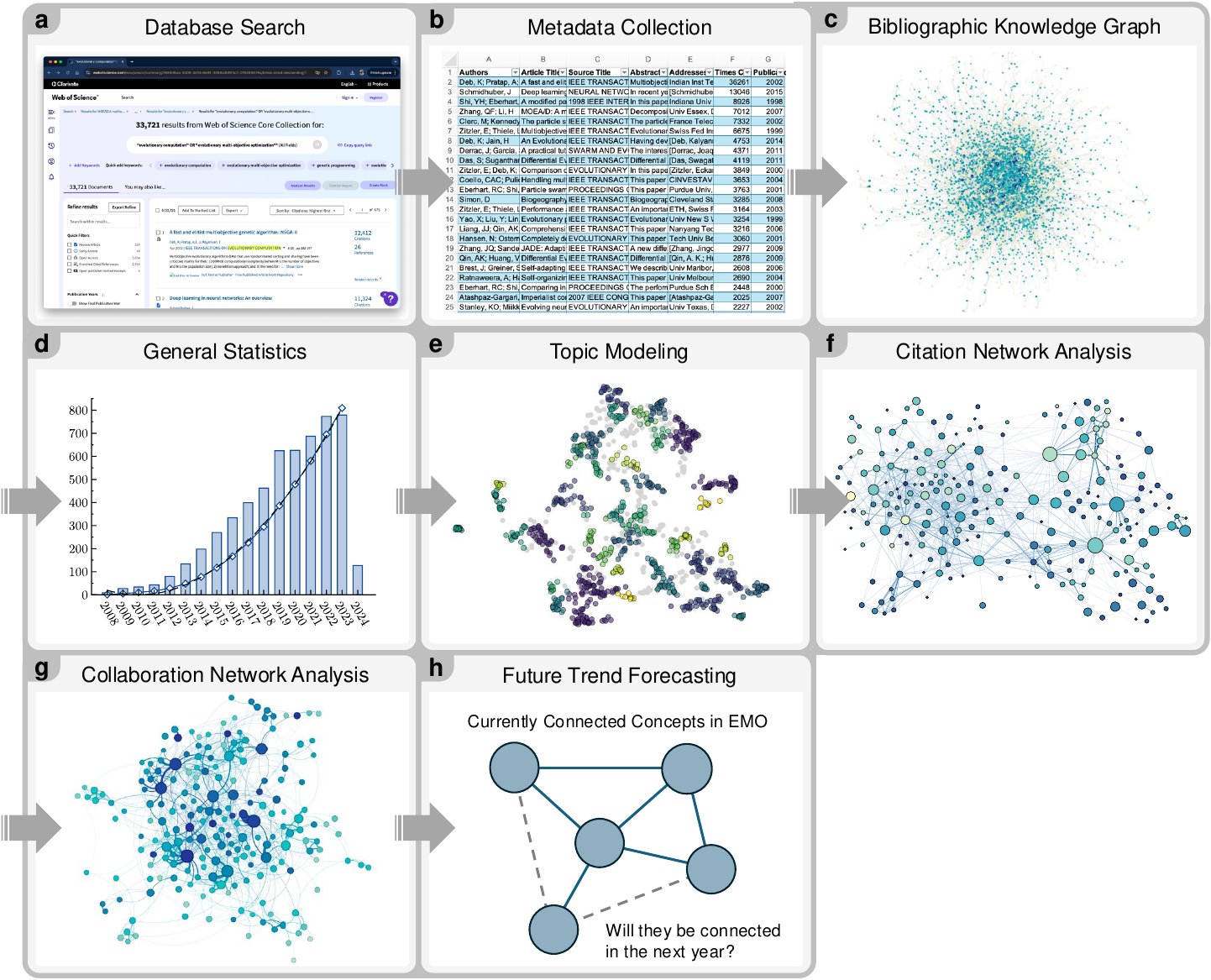}
    \caption{Overview of the \our\ workflow, featuring eight primary modules. \textbf{(a)} online literature database query for relevant literature. \textbf{(b)} collecting publication metadata from online databases. \textbf{(c)} constructing bibliographic knowledge graph that represents the literature landscape using collected metadata, wherein nodes are heterogenous entities like authors, papers, venues, etc., and edges indicate relationships among them. \textbf{(d)} general statistical analysis of the literature landscape, e.g., publications per year, geographical distributions of researchers. \textbf{(e)} topic modeling using paper embeddings and clustering algorithms. \textbf{(f)} citation network analysis that highlights the ``skeleton'' of existing research and characterizes the growth pattern of the research. \textbf{(g)} collaboration network analysis to reveal collaboration patterns in the community. \textbf{(h)} future trend forecasting by learning from hidden patterns embedded in the research landscape.}
    \label{fig:workflow}
\end{figure*}

In this work, we propose an end-to-end data-driven workflow, referred to as literature landscape analysis (\our) that systematically integrates, processes, and organizes diverse bibliographic metadata to build a structured, heterogeneous knowledge graph (see schematic overview in~\pref{fig:workflow}). It then applies a comprehensive suite of data mining tools---spanning descriptive statistics, data visualization, topic models, network science methods, and advanced machine learning algorithms---to uncover underlying research patterns and dynamics. By computationally synthesizing extensive high-dimensional literature data into meaningful insights, \our\ surpasses traditional, labor-intensive SLR methods in terms of scale, efficiency, and objectivity. Consequently, it better aligns with the COOT principles.

To demonstrate the effectiveness of \our, this paper focuses on decomposition-based evolutionary multi-objective optimization (EMO). Specifically, we use the multi-objective evolutionary algorithm based on decomposition (MOEA/D)~\cite{ZhangL07} as a case study. MOEA/D is a mainstream method in EMO, extensively researched across various domains, with over $6,000$ follow-up works.\footnote{Sourced from WoS. The number is lower than Google Scholar due to differences in inclusion criteria, such as excluding preprints, theses, and patents.} This volume of literature provides an ideal testbed for showcasing our workflow’s capabilities. Moreover, one of our authors has over $15$ years' experience in EMO and MOEA/D, which helps validate and interpret the data-driven findings. In summary, we collected and processed bibliographic data on more than $5,400$ papers, $10,000$ authors, $1,400$ venues, $78,000$ keywords, and $1,600$ institutions related to MOEA/D. We then applied \our\ to produce an \lq atlas\rq\ of the MOEA/D literature landscape. This paper is organized as follows.

\begin{itemize}
    \item \pref{sec:data} outlines our \textbf{data collection} protocol, specifying sources, selected publications, metadata, and methods for constructing the bibliographic knowledge graph.
    \item \pref{sec:geneal} provides a \textbf{general analysis} of MOEA/D-related publication trends, venues, authors, and disciplinary distributions, leveraging this knowledge graph.
    \item \pref{sec:topic} applies \textbf{topic modeling} to analyze key research themes and applications in MOEA/D, examining their spatial-temporal distributions and interconnections.
    \item \pref{sec:citation} explores the evolving landscape of MOEA/D research through \textbf{citation network} analysis, highlighting changes in research disruptiveness and using main path analysis to identify influential studies over time.
    \item \pref{sec:collaboration} investigates the \textbf{collaboration network} among MOEA/D researchers to uncover notable patterns.
    \item \pref{sec:prediction} presents a proof-of-concept on \textbf{forecasting} prospective research directions in the MOEA/D domain by learning from the existing research landscape.
    \item \pref{sec:interview} conducts a \textbf{subjective study} involving interviews with $20$ EMO researchers who found \our\ informative, validated its insights, and provided constructive feedback for further improvements.
    \item \pref{sec:conclusion} concludes this paper and sheds some light for future directions.
\end{itemize}

\section{Related Works}
\label{sec:related}

\subsection{Surveys on MOEA/D.} 

In recent years, there have been several survey papers on MOEA/D and decomposition-based EMO~\cite{Li24,XuXM20,TrivediSSG17,PinedaHDPSBM14,WangSLMGLM20,MaYLQZ20,Guo22a,XuXM19,MaYWJZ16}. They typically adopt a manual review approach. Although this strategy is small in scale, it provides valuable expert insights. Our work complements these expert-driven surveys by offering a broader, data-driven mapping of the entire MOEA/D literature landscape. In the sections that follow, we will compare some of our findings (e.g., prevalent topics identified by topic modeling) with those reported in these surveys. This comparison demonstrates how our workflow can uncover insights comparable to (or even broader than) expert reviews, yet at much lower cost in terms of time and required expertise.

\subsection{Data-driven-based Literature Exploration}

Data-driven exploration methods have appeared in several survey papers from other fields (e.g.,~\cite{LiuLLL13,HeldensHWMBN20,SunY17,JeongZD23}). However, these studies are often either \lq narrow\rq\ in breadth (focusing on one type of analysis) or \lq shallow\rq\ in depth (using simple or outdated techniques). For instance, \cite{LiuLLL13} and \cite{JeongZD23} restricted their reviews to citation network analysis, while~\cite{SunY17} relied solely on topic modeling. \cite{HeldensHWMBN20} applied both types of analysis, but they only reported basic network metrics and used non-negative matrix factorization for the topic model, which has been significantly outperformed by recent deep learning-based approaches~\cite{ChurchillS22}. In contrast, \our\ builds on a heterogeneous knowledge graph that integrates various types of bibliographic metadata. It then applies five different data-mining perspectives, each supported by state-of-the-art techniques. Combined, these perspectives provide a more comprehensive and in-depth understanding of the literature landscape than existing approaches.

\section{Data Collection}
\label{sec:data}
This section introduces our data collection protocol and the construction of the knowledge graph representing the MOEA/D literature landscape.

\subsection{Data Collection}

\subsubsection{Acquiring papers}
To identify research pertinent to MOEA/D, we initially included all papers citing Zhang and Li's seminal work~\cite{ZhangL07}. Our rationale is that such papers likely apply or extend MOEA/D, compare it with other EMO methods, or discuss it in a broader EMO context. We deemed all these papers—even those that only mention \cite{ZhangL07}—as valuable for accurately positioning MOEA/D within the EMO community. This initial filtering, conducted in March 2024, yielded $5,606$ works indexed by WoS. We chose WoS for its comprehensive coverage of over 225 million records from $34,000$ venues and $2.56$ billion citation relationships, as well as its well-curated metadata. We then refined the candidate pool using the following exclusion criteria, resulting in a final selection of $5,404$ papers published between 2007 and 2024.
\begin{itemize}
    \item[\ding{56}] Papers not written in English.
    \item[\ding{56}] Papers under four pages in length (e.g., short or work-in-progress papers such as GECCO Companion).
    \item[\ding{56}] Books, keynote records, unpublished manuscripts, workshop papers, and other non-peer-reviewed documents.
    \item[\ding{56}] Extended journal versions of conference papers.
\end{itemize}

\subsubsection{Collecting metadata}
The foundation of our data-driven analysis is a corpus of high-quality metadata for each paper. This metadata can be leveraged in data mining to reveal latent patterns. To support our study, we compiled $15$ meta-features from diverse sources, as shown in~\pref{tab:meta_data}. The WoS database itself provides a wealth of general features ($F_1$-$F_{12}$). Among the $5,404$ papers, only $113$ had missing fields, and these were excluded from the relevant analyses. We also inferred each paper's geographical origin ($F_{13}$) by parsing author addresses. To investigate authors' intent when citing MOEA/D, we extracted citation contexts ($F_{14}$) from each paper using the Semantic Scholar \texttt{S2ORC} dataset~\cite{LoWNKW20}. Finally, we prompted an LLM (\texttt{GPT-4o}) to process paper abstracts and extract EMO-related keywords ($F_{15}$), such as \lq Pareto Front\rq\ or \lq knee point\rq. We utilize these extracted keywords instead of author-provided terms, as they provide a more consistent and standardized framework for our analysis.

\begin{remark}{\faComment \, A Sample Citation Statement}
    \lq\lq \textit{These algorithms had applied into benchmark problems. Among all these variants only MOEA/D [1], MOEA/D-DD [4], MOEA/D-DU [8], MOEA/D-UR [16], and MOEA/D-URAW [17] had given the comparative results. For this reason, only the results belonging to these 5 algorithms has been reported on the paper} \dots\rq\rq \vspace{0.05cm} --- Section: Results
\end{remark}

\subsubsection{Constructing literature landscape}
We synthesized all publications and their metadata into a bibliographic knowledge graph that represents the MOEA/D literature landscape (see~\pref{def:landscape}). This graph includes five types of bibliographic keywords: $i)$ $5,404$ papers, $ii)$ $10,532$ researchers, $iii)$ $432$ venues, $iv)$ $78,490$ keywords, and $v)$ $1,661$ institutions. It also encodes multiple relationships among these nodes. For example, Zhang and Li (2007)~\cite{ZhangL07} \lq is cited by\rq\ Li \textit{et al.} (2015)~\cite{LiDZK15}, or Zhang and Li (2007)~\cite{ZhangL07} \lq is published in\rq\ \textit{IEEE Trans. Evol. Comput}. This knowledge graph contains all essential information for performing a comprehensive analysis of MOEA/D research.
\begin{definition}[Literature Landscape]
    \label{def:landscape}
    Let $\mathcal{P}$ be a set of publications relevant to a particular domain $\mathcal{D}$, and let $\mathcal{A}$, $\mathcal{I}$, $\mathcal{V}$, and $\mathcal{K}$ be sets of authors, institutions, venues, and keywords associated with these publications, respectively. We define a \emph{literature landscape} for $\mathcal{D}$ as a heterogeneous knowledge graph
    \[
    \mathcal{G} \;=\; (\mathcal{N}, \mathcal{E}),
    \]
    where
    \[
    \mathcal{N} \;=\; \mathcal{P} \,\cup\, \mathcal{A} \,\cup\, \mathcal{I} \,\cup\, \mathcal{V} \,\cup\, \mathcal{K}
    \]
    is the set of nodes and $\mathcal{E} \subseteq \mathcal{N}\times \mathcal{N}$ is the set of edges capturing relationships among nodes (e.g., \textit{author-of}, \textit{affiliated-with}, \textit{published-at}, \textit{mentions-keyword}, \textit{cites}). Each edge $(n_i, n_j)\in \mathcal{E}$ and node $n_i\in \mathcal{N}$ is associated with a type $\phi(n_i)\in \mathcal{T}$ and $\psi(n_i, n_j)\in \mathcal{R}$, respectively, where $\mathcal{T}$ and $\mathcal{R}$ are predefined sets of node and edge types, respectively. Edges and nodes can have additional attributes, such as frequency of co-authorship, number of citations, or citation counts of a publication.
\end{definition}

\begin{table}[t!]
    \centering
    \footnotesize
    \setlength{\defaultaddspace}{2pt}
    \caption{Paper meta-features considered in this survey.}
    \begin{tabular}{ccc}
        \hline
        \addlinespace
        ID       & Meta feature & Source \\ \hline
        \addlinespace
        $F_1$    & Authors & WoS \\ 
        $F_2$    & Institutions & WoS \\ 
        $F_3$    & Year & WoS \\ 
        $F_4$    & Title & WoS \\ 
        $F_5$    & Publication venue & WoS \\ 
        $F_6$    & Publication type (Journal or Conference) & WoS \\ 
        $F_7$    & Author keywords & WoS \\ 
        $F_8$    & Web of Science subject category & WoS \\ 
        $F_9$    & Publisher & WoS \\ 
        $F_{10}$ & Number of citations & WoS \\ 
        $F_{11}$ & Number of pages & WoS \\
        $F_{12}$ & References & WoS \\
        $F_{13}$ & Country & Eng. \\ 
        $F_{14}$ & Citation context & SS \\ 
        $F_{15}$ & Extracted EMO keywords & Eng. \\
        \addlinespace 
        \hline
        \addlinespace
        \multicolumn{3}{l}{\footnotesize Eng. implies engineered features. SS stands for Semantic Scholar.} \\
    \end{tabular}
    \label{tab:meta_data}
\end{table}

\section{General Data Analysis}
\label{sec:geneal}

We now start deciphering the topography of the MOEA/D research landscape using our constructed knowledge graph, starting from a general overview in this section.

\subsubsection{Publication trends}

\begin{figure}[t!]
    \centering
    \includegraphics[width=\linewidth]{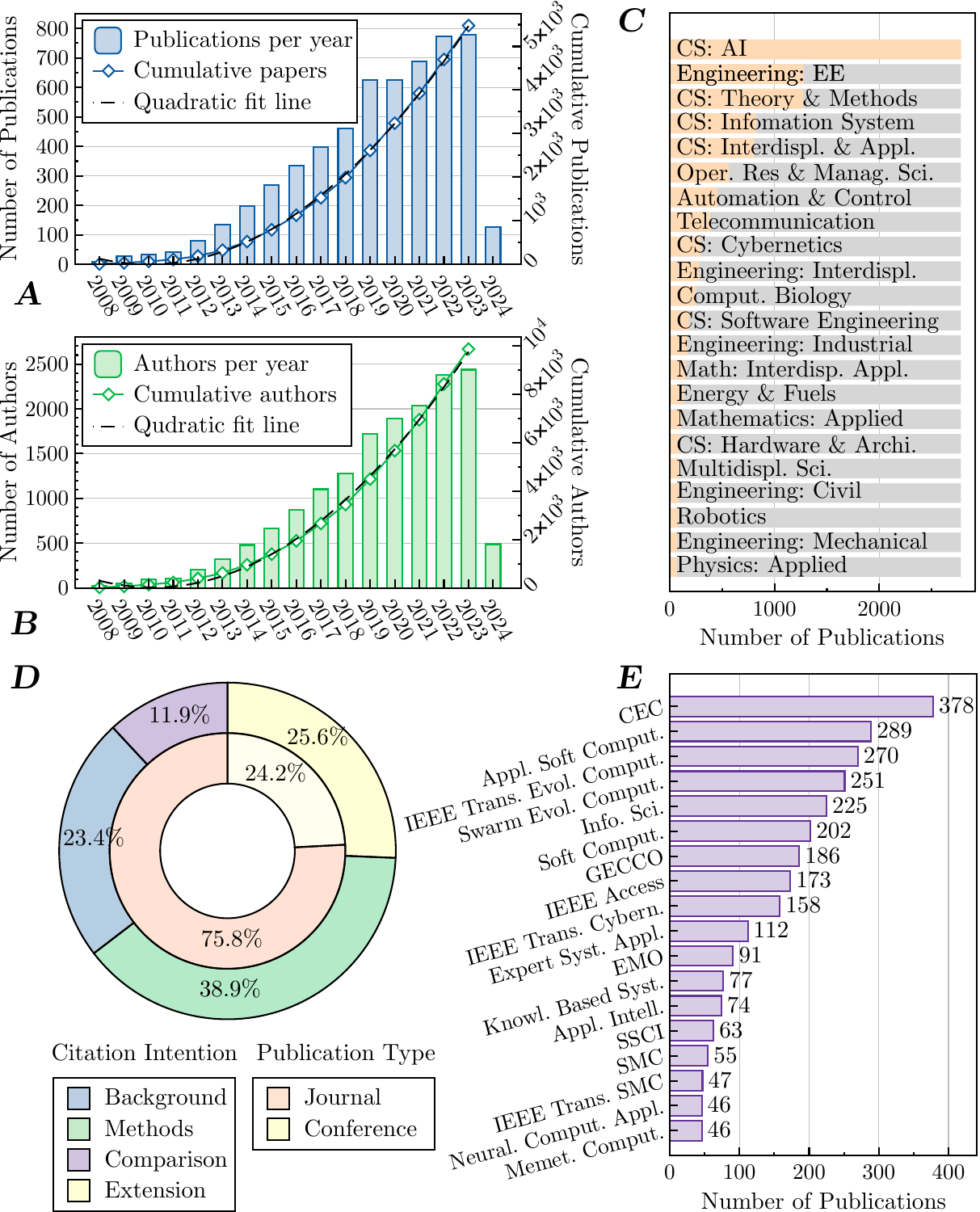}
    \caption{General information of surveyed MOEA/D literature. \textbf{(A)} Number of publications per year and its cumulative distribution. \textbf{(B)} Number of authors per year and its cumulative distribution. \textbf{(C)} Frequency of top-20 subject categories.  \textbf{(D)} Ring chart of pulication type (inner) and citation intention (outer). \textbf{(E)} Number of publications of the top-20 popular venues.}. 
    \label{fig:general_info}
\end{figure}

\begin{figure*}[t!]
    \centering
    \includegraphics[width=\linewidth]{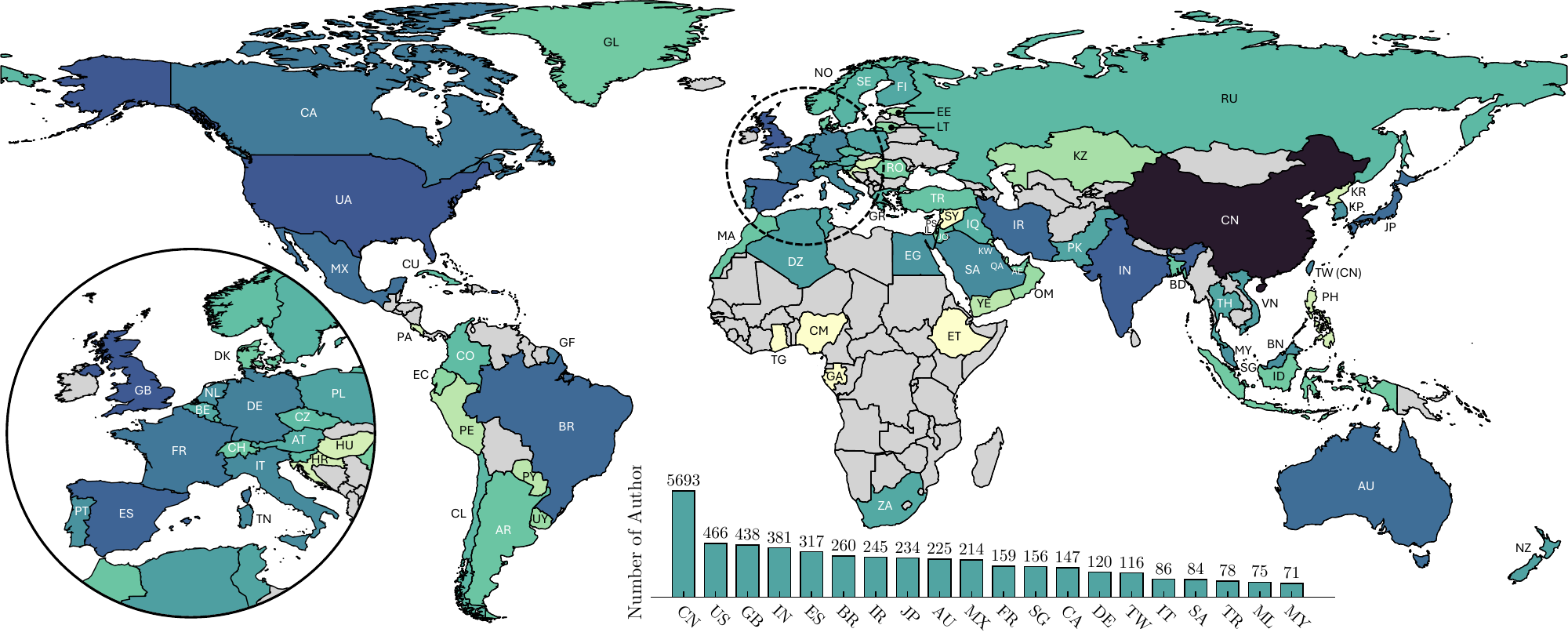}
    \caption{\textbf{(Top)} Geographic distribution of MOEA/D researchers. \textbf{(Bottom)} The number of researchers in the $20$ most active regions.}
    \label{fig:map}
\end{figure*}

The histogram in~\pref{fig:general_info}(A) illustrates the annual distribution of MOEA/D-related publications. The earliest follow-up studies appeared in 2008, and the publication count grew significantly thereafter, surpassing $100$ by 2013, $500$ by 2019, and nearly $800$ by 2023. On average, the field expanded at an annual rate of $40.5\%$. The cumulative number of publications, also shown in~\pref{fig:general_info}(A), closely follows a quadratic growth curve ($R^2 = 0.999$). These trends suggest that MOEA/D research will likely continue its robust growth, with more papers expected to further advance the field.

\subsubsection{Researcher involvement}
\pref{fig:general_info}(B) shows the annual count of authors involved in MOEA/D publications, along with its cumulative distribution. This pattern echoes the trends observed in~\pref{fig:general_info}(A). Notably, since 2017, over $1,000$ researchers have contributed to MOEA/D research each year, and the total number of researchers surpassed $10,000$ in early 2024. The cumulative distribution also follows a quadratic trend ($R^2=0.998$), suggesting a continued influx of new researchers to this community.

\subsubsection{Publication venues}
\pref{fig:general_info}(E) shows the most common venues for MOEA/D publications, while the inner ring of~\pref{fig:general_info}(D) illustrates the distribution by type. Journals are the primary publication venue, accounting for $76\%$ of all publications. Among these, \href{https://www.sciencedirect.com/journal/applied-soft-computing}{Appl. Soft. Comput}, \href{https://ieeexplore.ieee.org/xpl/RecentIssue.jsp?punumber=4235}{IEEE Trans. Evol. Comput.}, and \href{https://www.sciencedirect.com/journal/swarm-and-evolutionary-computation}{Swarm. Evol. Comput.} lead in publication counts. Conferences and symposiums represent the remaining $24\%$, led by \href{https://2024.ieeewcci.org/}{CEC}, \href{https://gecco-2024.sigevo.org/HomePage}{GECCO}, and \href{https://emo2025.org/}{EMO}. Although \href{https://direct.mit.edu/evco}{Evol. Comput.} ($26$ papers) and \href{https://link.springer.com/conference/ppsn}{PPSN} ($39$ papers) are well recognized publication venues, their lower publication counts place them outside the top ranks.

Beyond evolutionary computation, MOEA/D research appears in other computational intelligence venues, such as \href{https://cis.ieee.org/publications/t-fuzzy-systems}{IEEE Trans. Fuzzy Syst.}, \href{https://ieee-itss.org/pub/t-its/}{IEEE Trans. Intell. Transp. Syst.}, \href{https://www.grss-ieee.org/publications/transactions-on-geoscience-remote-sensing/}{IEEE Trans. Geosci. Remote Sens.}, and \href{https://cis.ieee.org/publications/t-neural-networks-and-learning-systems}{IEEE Trans. Neural Netw. Learn. Syst.}. This demonstrates the interdisciplinary nature of the MOEA/D literature landscape. For a broader view of domain diversity, \pref{fig:general_info}(C) categorizes publications across various subject areas in WoS. While these categories are coarse-grained, a more detailed topic modeling approach is discussed in~\pref{sec:topic}.

\subsubsection{Geographical regions}
\pref{fig:map} illustrates the geographical distribution of MOEA/D researchers. China leads in MOEA/D activity, accounting for over half of all researchers in this domain. The United States, the United Kingdom, India, and Spain also contribute substantially. This pattern highlights a global interest in MOEA/D research, spanning $82$ countries across all major continents.

\subsubsection{Citation intents}
By analyzing citation statements ($F_{14}$) from~\pref{sec:data}, we identified the underlying intent of each MOEA/D citation, provided that the paper offered a relevant citation statement. Although citation intent classification has been explored in earlier work (e.g., \cite{CohanAZC19}), only recent advances in LLMs have substantially improved performance \cite{KunnathPK23}. Here we used \texttt{GPT-4o} to annotate $4,675$ citation statements into four categories, shown in the outer ring of \pref{fig:general_info}:
\begin{itemize}
    \item \underline{Background}: About $25\%$ mention MOEA/D in a broader context, offering historical detail, justifications, or other supporting information.
    \item \underline{Method}: Nearly $40\%$ refer to MOEA/D as the primary methodology, often applying EMO in various disciplines.
    \item \underline{Extension}: Around $26\%$ extend MOEA/D by introducing new features or adaptations to the original framework.
    \item \underline{Comparison}: The remaining $12\%$ benchmark new methods against MOEA/D or evaluate its performance relative to other approaches.
\end{itemize}


 

\section{Topic Modeling}
\label{sec:topic}

\begin{figure*}[t!]
    \centering
    \includegraphics[width=.825\linewidth]{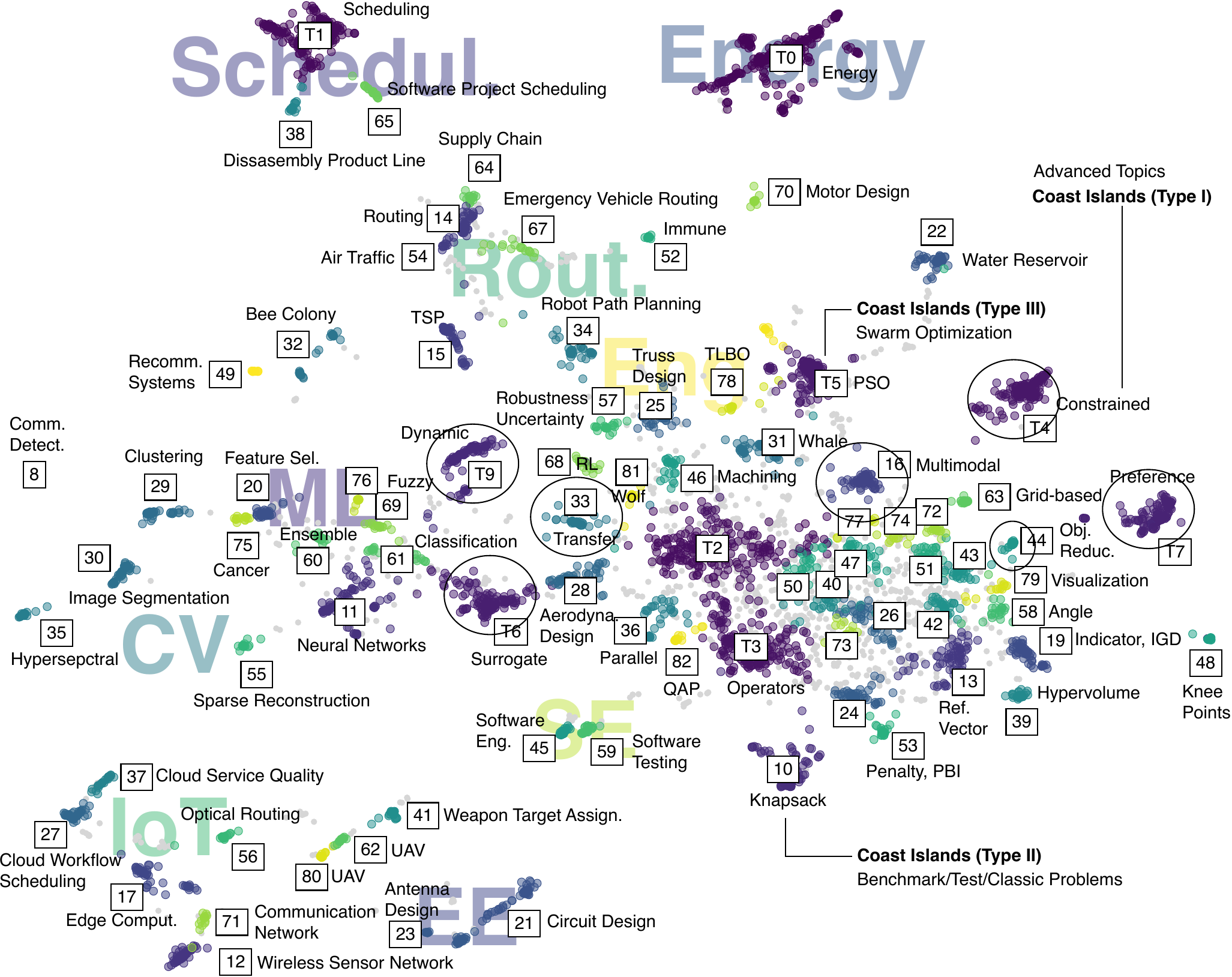}
    \caption{Low-dimensional visualization of the MOEA/D literature landscape by projecting the paper embeddings using UMAP. Papers are colored and labeled by \texttt{BERTopic} topics. Outliers are shown in light gray.}
    \label{fig:topic_2d}
\end{figure*}


To uncover prevalent research trends in MOEA/D from our knowledge graph, we applied topic modeling~\cite{ChurchillS22} to cluster papers by analyzing their titles and abstracts. These sections usually offer concise summaries of a paper’s focus and contributions, making them suitable for extracting thematic insights. We followed the pipeline from~\cite{Grootendorst22}, which learns a high-dimensional representation of each paper in a latent embedding space. Papers addressing similar topics lie close to each other in this space. We used the \texttt{voyage-2-large} model from \href{https://www.voyageai.com/}{VOYAGE~AI} for embedding, as it is a top-performing text model. Next, we applied \texttt{UMAP}~\cite{McInnesHSG18} to reduce the embeddings to five dimensions and mitigate the curse of dimensionality~\cite{AggarwalHK01}. Finally, we used \texttt{HDBSCAN}~\cite{McInnesHA17} to cluster the papers and identify distinct research topics. This pipeline captures nuanced semantic similarities and often achieves better coherence, diversity, and scalability than traditional probabilistic models~\cite{AbdelrazekEGMH23,BleiNJ03}.

We then assigned a human-readable topic label to each cluster using a class-based variant of TF-IDF~\cite{Joachims97}. In practice, some interdisciplinary topics (e.g., \lq hyperspectral imaging\rq) remain challenging to interpret, even with TF-IDF keywords. To address this, we used \texttt{GPT-4o} to generate concise summaries by providing TF-IDF outputs, paper titles, and exemplar abstracts. Throughout this process, we iteratively refined model selection and hyperparameter settings with feedback from an EMO research group comprising over $15$ postgraduate students and $4$ professors.

In total, we identified $83$ distinct topics and ranked them by their size. The largest cluster (\texttt{T0}) includes over $1,200$ papers, while the smallest (\texttt{T82}) contains $50$. Word clouds of these topics appear in Figure A1 and A2 of Appendix B\footnote{Appendix is available at: https://zenodo.org/records/15088252.}, along with \texttt{GPT-4o} summaries for each. We also performed hierarchical clustering on the $83$ topics to form a tree structure (see Figure A3 of Appendix B). Broadly, MOEA/D research splits into two main streams: methodological enhancements and extensions ($40$ topics) and application-driven research ($43$ topics). The former focuses on refining and advancing MOEA/D, whereas the latter addresses applications across various domains. Each stream is further subdivided into distinct themes.

\begin{remark}{\faGavel \, Remark: \textit{outliers}}
    Not all papers were assigned to a topic. Such papers are treated as outliers and shown in light gray in~\pref{fig:topic_2d}. An outlier is not necessarily irrelevant; it may occupy a mid-region between two or more topics in the semantic space. Although overlapping clustering methods can reduce outliers, we allowed a reasonable amount here to maintain clear topic boundaries.
\end{remark}

\begin{remark}{\faGavel \, Remark: \textit{topic size}}
    The size of the topics here should be interpreted with caution. First, the presence of outliers skews absolute counts, so relative comparisons are more meaningful. Second, since this paper originates from EMO research, we tuned \texttt{BERTopic} to highlight methodological topics while retaining only a high-level view of application topics. This emphasis explains why \texttt{T0} and \texttt{T1} are larger than \texttt{T2} and \texttt{T3}.
\end{remark}

We further visualized the MOEA/D literature landscape via a two-dimensional \texttt{UMAP} projection in~\pref{fig:topic_2d}. This map preserves small distances between topically similar papers, offering a broader context for the identified clusters. Methodological topics appear in the central \lq continent\rq, with specialized areas (e.g., constrained multi-objective optimization, surrogate modeling) forming distinct \lq coast islands\rq\ surrounding the continent. Scheduling and routing problems occupy the upper region, while electrical engineering and computer science applications appear in the lower and left regions, respectively. In the following sections, we delve deeper into this topographic overview.

\vspace{-1em}
\subsection{Methodological Topics}

The methodological topics focus on the broader family of MOEAs. We classify them into five main themes, each spotlighting distinct research directions and problem types.

\subsubsection{MOEA/D} Many follow-up studies on MOEA/D fall under \texttt{T2}~\cite{LiZKLW14,AsafuddoulaRS15,YuanXWZY16}, which is the largest topic in this methodological group. While \texttt{T2} is more like a general one, we noticed that several topics mentioned in the latest MOEA/D survey~\cite{Li24} appear as separate clusters, e.g., weight vector settings (\texttt{T24}), archives (\texttt{T47}), estimation of distribution methods (\texttt{T50}), and penalties (\texttt{T53}).

\subsubsection{General MOEA} This theme covers topics relevant to the wider EMO community, such as performance indicators (\texttt{T19}, \texttt{T39}), parallelization (\texttt{T36}), knee points (\texttt{T48}), robust optimization (\texttt{T57}), and visualization (\texttt{T79}). We also identified clusters on dominance relationships (\texttt{T43}) and balancing convergence and diversity (\texttt{T42}). Several groups focus on NSGA-II~\cite{DebAPM02} or NSGA-III~\cite{DebJ14}, often citing MOEA/D for comparison. Their inclusion supports our paper selection protocol in~\pref{sec:data}, since \texttt{BERTopic} naturally recognizes isolated or unrelated groups. Finally, \texttt{T3} addresses general MOEA research. It includes surveys, guidance on MOEAs, and software packages like \texttt{Pymoo}~\cite{BlankD20} and \texttt{PlatEMO}~\cite{TianCZJ17}.

\subsubsection{Advanced topics} This theme includes constrained MO (\texttt{T4}), surrogate modeling (\texttt{T6}), preference-based and interactive MO (\texttt{T7}, \texttt{T13}), and dynamic MO (\texttt{T9}). We also observed emerging areas like multi-modal MO (\texttt{T16}), transfer learning and multi-tasking (\texttt{T33}), large-scale MO (\texttt{T40}), and objective reduction (\texttt{T44}).

\subsubsection{Swarm intelligence} Swarm-based methods represent another active line of research. Particle swarm optimization appears in \texttt{T5}~\cite{PengZ08,MoubayedPM10,MartinezC11}, while whale optimization (\texttt{T31})~\cite{Abdel-BassetMM21}, artificial ant/bee colony (\texttt{T32})~\cite{KeZB13,ZhangXLLS20,NingZSF20,WuQJZX20}, immune algorithms (\texttt{T52})~\cite{ShangJLM12}, and grey wolf optimizers (\texttt{T81})~\cite{MirjaliliSMC16} form separate subtopics.

\subsubsection{Benchmark test problems} Classic MOPs make up this final methodological theme. The multi-objective knapsack problem (\texttt{T10}, e.g.,~\cite{IshibuchiAN15}) is most prominent, followed by multi-objective TSP (\texttt{T15}, e.g.,~\cite{ZhouGZ13}). We also identified a dedicated cluster for benchmarking and test problems (\texttt{T26}, e.g.,~\cite{ChengJOS17}), as well as smaller groups focusing on quadratic assignment problems (\texttt{T82}, e.g.,~\cite{VenskeALD22}).

\vspace{-1em}
\subsection{Applications}

In addition to methodological advances, our topic modeling reveals a broad range of MOEA/D applications. These applications fall into three main categories.

\subsubsection{Operations research} Many MOEA/D applications involve planning, scheduling, or routing tasks within operations research (OR). For instance, \textit{planning \& scheduling} covers flow-shop (\texttt{T1}, e.g.,~\cite{SenguptaDNVP12,ZhouLZG17,ChangCZL08}) and supply-chain (\texttt{T64}) scheduling, portfolio selection in finance (\texttt{T18}, e.g.,~\cite{MishraPM14}), software project scheduling in software engineering (\texttt{T65}, e.g.,~\cite{XiangYZH20}), and the weapon target assignment problem in military OR (\texttt{T41}, e.g.,~\cite{Li0XD15}). Meanwhile, \textit{routing problems} (\texttt{T34}, \texttt{T14}, \texttt{T62}, \texttt{T54}, \texttt{T66}, e.g.,~\cite{QiHLHL15}) cover robotics, cars, UAVs, aircraft, and satellites.

\subsubsection{Engineering} MOEA/D is also widely applied in real-world engineering settings. Examples include designing analog circuits (\texttt{T21}) and antennas (\texttt{T23}, e.g.,~\cite{DingW13}) in electronics engineering, optimizing wireless sensors (\texttt{T12}, \texttt{T17}) and communication networks (\texttt{T56}, \texttt{T71}, e.g.,~\cite{XuDQL18,KonstantinidisY11,XingWLLQ17}) in communication engineering, and solving a variety of engineering shape design problems (\texttt{T25}, \texttt{T28}, \texttt{T46}, \texttt{T70}, e.g.,~\cite{Ho-HuuHVC18,ChampasakPPBY20,KumarJTM22a}).

\subsubsection{Computer science} In the last decade, computer science (CS) has also experienced a surge in MOEA/D applications, particularly in machine learning (ML). Prominent examples include multi-objective neural architecture search (\texttt{T11}, e.g.,~\cite{JiangHLWLH20}) and multi-objective feature selection (\texttt{T20}, e.g.,~\cite{JiaoNXZ23}). Other ML tasks, such as clustering (\texttt{T29}, e.g.,~\cite{MukhopadhyayMB}), recommendation (\texttt{T49}, e.g.,~\cite{CaoZLY21}), and reinforcement learning (\texttt{T68}, e.g.,~\cite{LiuXH15}), also benefit from MO. 

\subsection{Topic distribution over time}

\begin{figure}[t!]
    \centering
    \includegraphics[width=\linewidth]{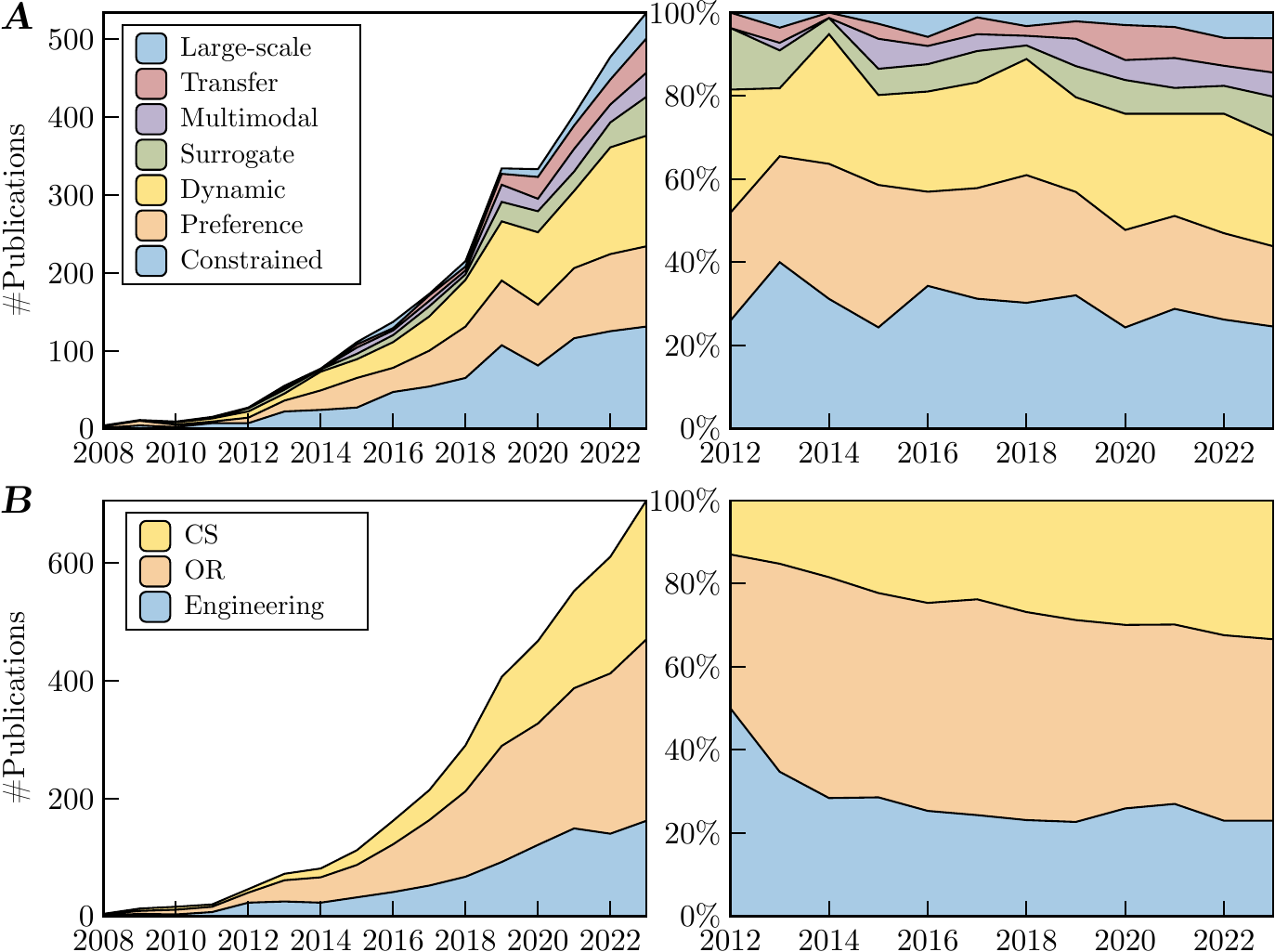}
    \caption{\textbf{(Left)} Number of publications per year and \textbf{(Right)} relative percentages for \textbf{(A)} the $7$ topics on MOs variants, \textbf{(B)} the $3$ application domains. For the right panel, we truncated the time range to 2012-2023 as the number of publications before 2012 is relatively small.}
    \label{fig:trend1}
\end{figure}

\begin{figure}[t!]
    \centering
    \includegraphics[width=\linewidth]{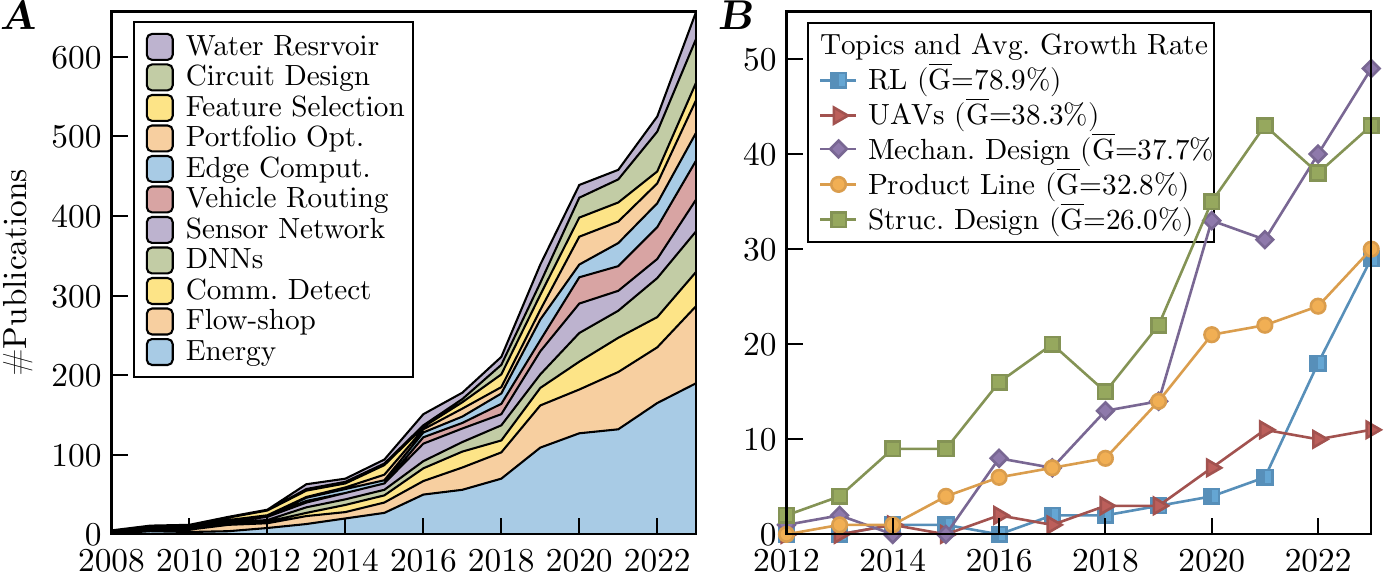}
    \caption{Number of publications per year for \textbf{(A)} top-10 popular applications topics and \textbf{(B)} top-5 emerging application topics.}
    \label{fig:trend2}
\end{figure}

In addition to examining the spatial distribution of topics, we also investigated how research interests evolve over time. First, we refined the paper-to-topic assignments by manually crafting query strings based on paper abstracts. This approach allows papers to belong to multiple topics, increasing the granularity of our analysis. We then applied the method from~\cite{Rosen-ZviGSS04} to model topic distributions over time, focusing especially on seven advanced variants of MO and MOEA/D applications.

\subsubsection{Advanced research topics}
\pref{fig:trend1} shows the growth of these seven advanced topics from 2008 to 2023. Overall, they follow the upward publication trend seen in \pref{fig:general_info}. Constrained MOPs (\texttt{T4}), preference-based MOPs (\texttt{T7}, \texttt{T13}, \texttt{T48}), and dynamic MOPs (\texttt{T9}) stand out (\pref{fig:trend1}A), accounting for about $400$ papers in 2023. Even after adjusting for overlap, they represent $319$ unique publications—nearly $40\%$ of the year's total (about $800$). Although these topics dominate in absolute terms, their share has declined because of new areas like surrogate-assisted MO (\texttt{T6}), multi-modal MO (\texttt{T16}), multi-task MO (\texttt{T33}), and large-scale MO (\texttt{T40}, \texttt{T44}). This indicates a diversification of research interests.

\subsubsection{General trend on applications}
We next examine \pref{fig:trend1}(B), which tracks the evolution of three broad MOEA/D application domains. All domains grew steadily, but growth in CS accelerated after 2012. This is in line with the rapid development of ML in the past decade. Engineering-focused applications saw a relative decline, while OR maintained a consistent share. However, many OR-related topics also stem from engineering, suggesting that engineering usage has not necessarily shrunk. Instead, the increasing diversity of MOEA/D applications may obscure its continued presence in traditional engineering.

\subsubsection{Top-$10$ application topics}
\pref{fig:trend2}(A) highlights the top-$10$ application topics of MOEA/D. Energy (\texttt{T0}), flow-shop scheduling (\texttt{T1}), and community detection (\texttt{T8}) lead this list. Notably, energy applications have surged in the past decade, reflecting global efforts to meet the United Nations' \href{https://worldtop20.org/global-movement/?gad_source=1&gclid=Cj0KCQjw-r-vBhC-ARIsAGgUO2A-O5LU_0WjkciDMK94g9q5-3FDdn7neZ6qzteP6L06QZ2uPGk6eioaAoScEALw_wcB}{sustainable development goals}. Other popular topics include neural network design (\texttt{T11}), wireless sensor network optimization (\texttt{T12}), vehicle routing (\texttt{T14}), edge computing (\texttt{T17}), portfolio selection (\texttt{T18}), feature selection (\texttt{T20}), analog circuit design (\texttt{T21}), and reservoir management (\texttt{T22}).

\subsubsection{Emerging application topics}
We also identified five rapidly growing application topics, selected by their annual publication growth since 2018. Although some high-growth topics from \pref{fig:trend2}(A) reappear, we focus here on those that are relatively new to the community. As shown in \pref{fig:trend2}(B), reinforcement learning (RL, \texttt{T68}) has expanded by nearly fourfold in the past two years. MOEA/D applications in UAVs (\texttt{T62}) rose significantly after 2020, coinciding with the maturation and commercialization of UAV technologies. Traditional engineering areas also appeared, such as structural design (\texttt{T25}) and mechanical design (\texttt{T46}), as well as product line scheduling (\texttt{T38}), all of which have recorded notable publication increases in recent years.

Finally, several topics in \pref{fig:trend1}(A) and \pref{fig:trend2}(A) show a \lq pause\rq\ or dip around 2020, likely due to COVID-19 disruptions. Nonetheless, publication growth rebounded strongly in 2022. Interestingly, this slowdown is less visible in the broader trends of \pref{fig:trend1}(B). The presence of many diverse topics may have buffered the overall MOEA/D research landscape against sharper fluctuations in specific areas.

\vspace{-1.0em}
\subsection{Topic Linkages}

\begin{figure}[t!]
    \centering
    \includegraphics[width=\linewidth]{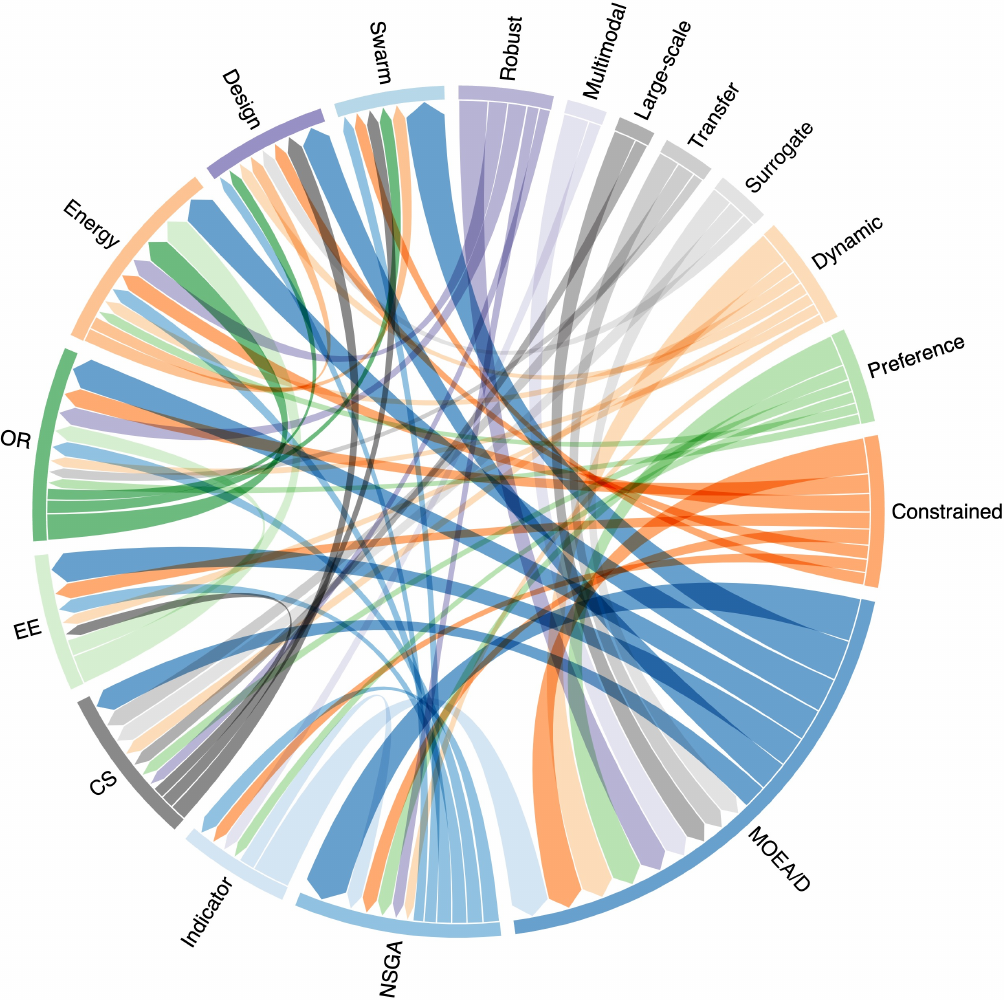}
    \caption{Chord diagram showing the linkages between different sectors of research themes in the MOEA/D landscape. This is determined via keyword co-occurrence in paper abstracts. A thicker chord indicates a stronger linkage between two topics. Direction of the linkages is only for enhanced readability.}
    \label{fig:topic_linkage}
\end{figure}

Topic modeling also offers a clear view of how major research themes in the MOEA/D landscape interconnect. We present these linkages as a chord diagram in~\pref{fig:topic_linkage}. The general MOEA/D theme\hspace{-0.2em}\coloredsquare{set_blue}\hspace{-0.2em} shows strong connections with nearly all other topics, including both methodological studies and real-world applications. This illustrates MOEA/D's broad impact across diverse areas. Moving counterclockwise, we see the seven advanced MO variants, each linked primarily to specific applications as well as the core MOEA/D theme.

Constrained MO\hspace{-0.2em}\coloredsquare{set_orange}\hspace{-0.2em} is the most prominent among these variants, drawing attention from every major application domain. Preference-based MO\hspace{-0.2em}\coloredsquare{set_green_light}\hspace{-0.2em} features heavily in CS\hspace{-0.2em}\coloredsquare{set_grey}\hspace{-0.2em}, OR\hspace{-0.2em}\coloredsquare{set_green}\hspace{-0.2em}, and power engineering\hspace{-0.2em}\coloredsquare{set_mid_orange}\hspace{-0.2em}, indicating that its relative share in other domains falls below the threshold $\epsilon=0.15$. Dynamic MO\hspace{-0.2em}\coloredsquare{set_light_orange}\hspace{-0.2em} also has robust links with each application field, reflecting its ongoing growth. Among the other four MO variants, most show strong ties to CS, except for multi-modal MO\hspace{-0.2em}\coloredsquare{set_light_purple}\hspace{-0.2em}. Notably, multi-task MO\hspace{-0.2em}\coloredsquare{set_mid_grey}\hspace{-0.2em} is prominent in OR\hspace{-0.2em}\coloredsquare{set_green}\hspace{-0.2em}, while many engineering design\hspace{-0.2em}\coloredsquare{set_purple}\hspace{-0.2em} tasks adopt surrogate modeling\hspace{-0.2em}\coloredsquare{set_light_grey}\hspace{-0.2em}.

We also observe overlaps between application domains. Many power engineering\hspace{-0.2em}\coloredsquare{set_mid_orange}\hspace{-0.2em} studies address planning and scheduling, which are traditionally part of OR\hspace{-0.2em}\coloredsquare{set_green}\hspace{-0.2em}. Power engineering strongly intersects with broader electrical engineering (EE)\hspace{-0.2em}\coloredsquare{set_green_mid}\hspace{-0.2em}, as shown by a large chord linking the two. NSGA and swarm optimizers appear across many topics but at lower connection strengths, reflecting this analysis’s focus on MOEA/D. Performance indicators\hspace{-0.2em}\coloredsquare{set_light_blue}\hspace{-0.2em} in MO mainly link to methodological themes. Finally, robustness and uncertainty in MO\hspace{-0.2em}\coloredsquare{set_mid_purple}\hspace{-0.2em} draw interest from both methodological and application-oriented research.

\section{Citation Network Analysis}
\label{sec:citation}

While topic modeling in~\pref{sec:topic} offers a unique high-level view of the MOEA/D literature landscape, it treats each paper as an isolated entity. In contrast, scientific publications are inherently connected through citation networks, which serve as channels for knowledge accumulation and dissemination. Citation network analysis~\cite{Price65} has been widely used to understand research themes, trace field evolution, and identify or predict influential works~\cite{KuhnPH14,FortunatoBBEHMMRSUVWWB18}. This section explores the citation relationships in our knowledge graph from three perspectives to shed new light on the MOEA/D literature landscape.

\begin{figure}[t!]
    \centering
    \includegraphics[width=\linewidth]{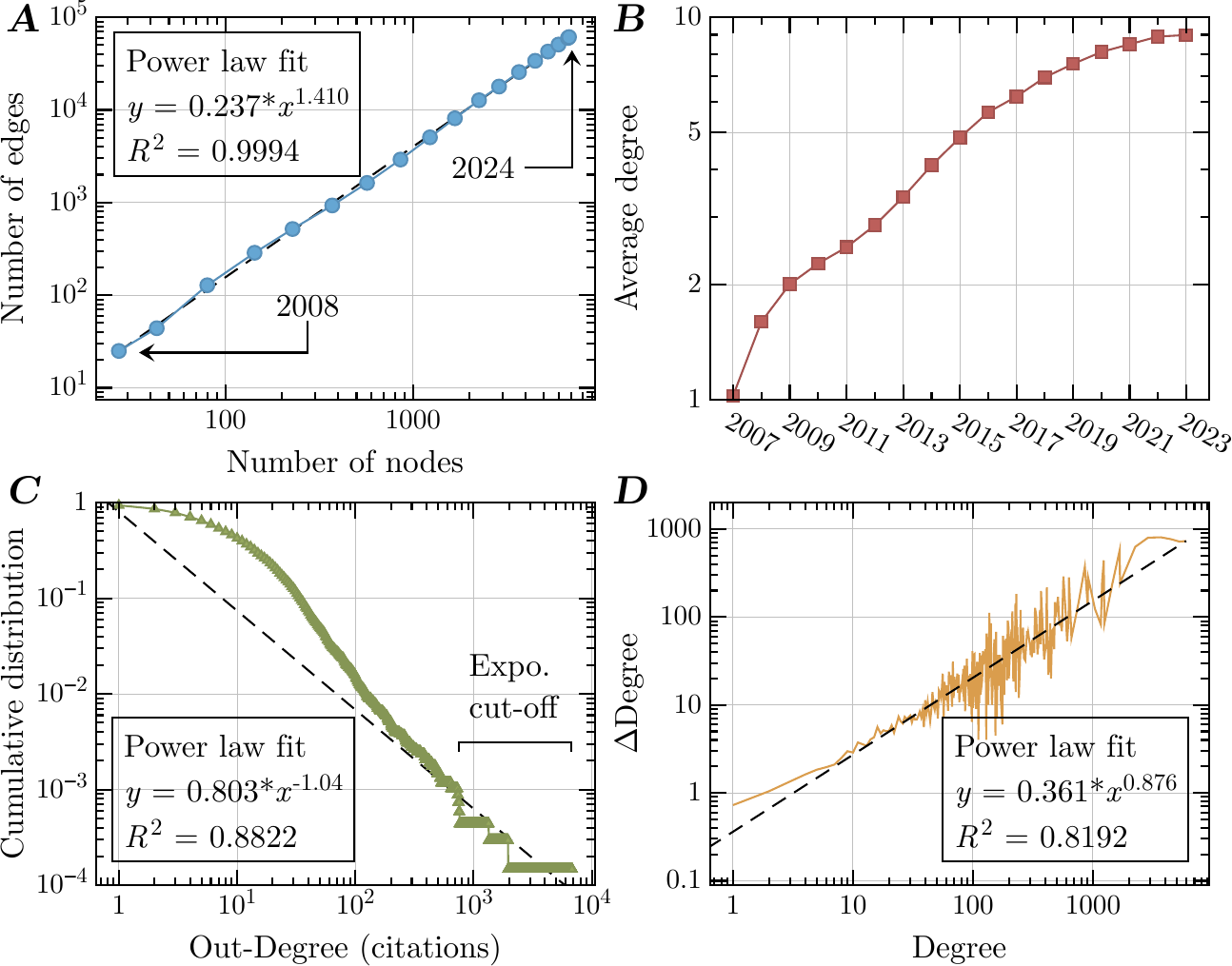}
    \caption{General patterns for MOEA/D citation networks. \textbf{(A)} The growth of the network nodes and edges from 2008 to 2024. \textbf{(B)} The average degree of the network each year. \textbf{(C)} Cumulative distribution of the number of citations of each paper within the network. \textbf{(D)} The number of new citations of a paper per year as a function of the citation it has already collected.}
    \label{fig:cn_metrics}
\end{figure}

\subsection{Network Evolution}

The MOEA/D citation network has grown significantly over the past decade. As~\pref{fig:cn_metrics}(A) shows, this expansion follows the \textit{densification power law}~\cite{LeskovecKF05}, with exponent $\alpha=1.41$. This indicates faster-than-linear growth ($\alpha=1$) and leads to a rising average out-degree (expected citations within the network), as shown in~\pref{fig:cn_metrics}(B).

Despite increased network density, not all papers receive citations equally. \pref{fig:cn_metrics}(C) reveals that the network's citation\footnote{Throughout this section, the citation index is the out-degree of each node in our MOEA/D citation network and may be lower than a paper’s total citation count.} distribution is highly skewed. It follows a power law with exponent $\alpha=1.04$, classifying the network as \textit{scale-free}~\cite{BarabasiA99}. Although the average citation count rises over time, only about $6\%$ of papers have more than $10$ citations by March 2024, just $1\%$ exceed $100$ citations, and only a handful surpass $1,000$. These findings mirror patterns in other scientific fields~\cite{Redner98} and suggest three main factors that drive citation polarization.

\paragraph{Preferential citations} A well-known model for understanding scale-free networks is \textit{preferential attachment}~\cite{BarabasiA99}, where newly added nodes tend to connect with those that already have a high degree. In our network, \pref{fig:cn_metrics}(D) shows that annual citation increases correlate with a paper's existing citation count, yielding a power law exponent of $\alpha=0.876$. This bias, or \lq rich-get-richer\rq\ dynamic, underpins the MOEA/D citation network's scale-free structure.

\paragraph{Knowledge obsolescence} Citation growth can be dampened by \lq knowledge obsolescence\rq, where older works attract fewer new citations over time~\cite{WangSB13}. \pref{fig:cn_metrics}(C) indicates a drop-off for papers cited more than $1,000$ times, suggesting an exponential cut-off~\cite{AmaralSBS00} and signifying that even highly cited works eventually lose momentum.

\paragraph{Competition} A third factor involves competition among papers on related topics, governed by a \textit{fitness} score~\cite{BianconB01}. This fitness reflects a paper's relevance and appeal to the community, thus influencing the distribution of citations within similar research areas.

\subsection{Paper Disruptiveness}

\begin{figure}[t!]
    \centering
    \includegraphics[width=\linewidth]{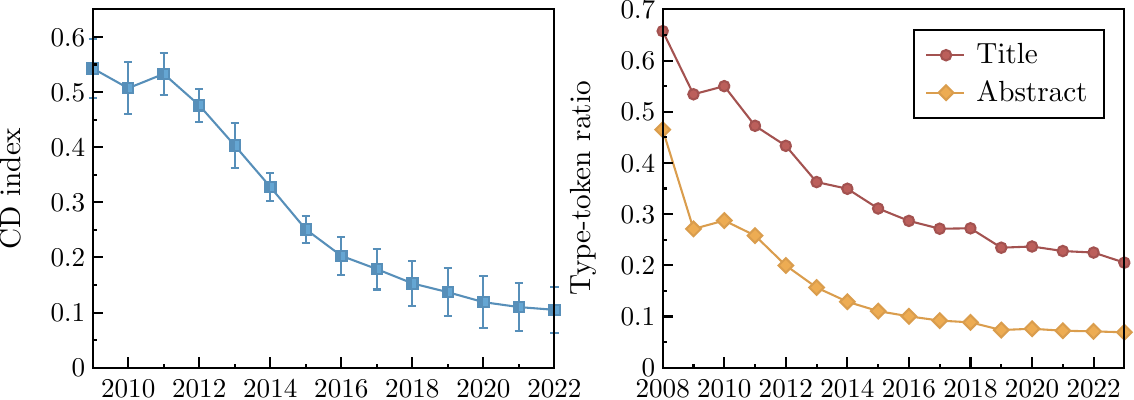}
    \caption{Paper disruptiveness for MOEA/D literature across time as measured by \textbf{(Left)} the CD index \textbf{(Right)} type-token ratio in paper titles and abstracts. Note that as the calculation of CD index requires information regarding both ancestor and successor papers, the data ranges from 2009 to 2022.}
    \label{fig:disruptiveness}
\end{figure}

Although MOEA/D literature has grown substantially, more papers do not necessarily imply more disruptive knowledge. Many works may echo or refine established ideas rather than introducing genuinely new paradigms. To assess the disruptive nature of these publications, we use the CD index, a metric from the scientometrics community~\cite{FunkO17}. It evaluates a paper's influence by comparing its citations with those of its predecessors. Papers with a CD index near $1$ are considered disruptive, because subsequent citing works tend not to cite earlier sources. Conversely, a CD index near $-1$ signifies consolidation: later citations also reference the paper's predecessors, thus reinforcing existing knowledge. By applying the CD index, we can distinguish whether MOEA/D literature consolidates prior research or fosters significant innovation.

As shown on the left of~\pref{fig:disruptiveness}, the disruptiveness of MOEA/D publications has declined over time. The CD index fell from $0.54$ in 2009 to $0.10$ in 2022. This observation aligns with linguistic analysis, which finds that fewer new words appear in paper abstracts and titles. The resulting drop in type-token ratio~\cite{ParkLF23}, depicted in~\pref{fig:disruptiveness} (Right), suggests diminishing novelty in language use. Such patterns mirror broader trends in scientific literature and may stem from a shortage of \lq low-hanging fruit\rq~\cite{ParkLF23,ChuE21}.

\subsection{Main Path Analysis}

\begin{figure}[t!]
    \centering
    \includegraphics[width=\linewidth]{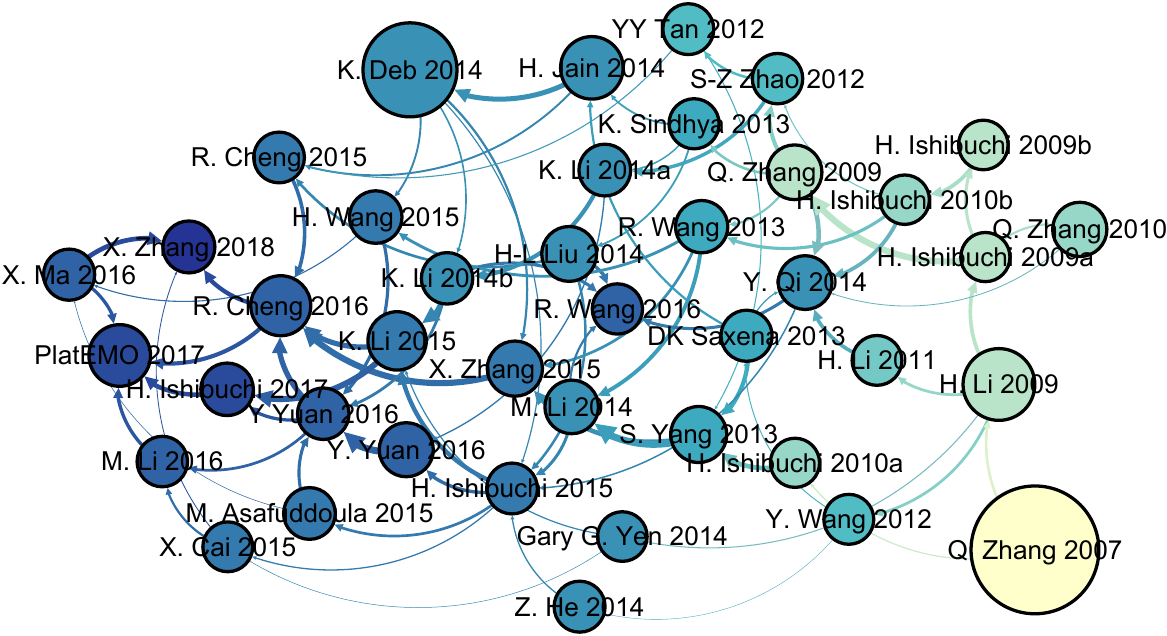}
    \caption{The most essential 40 nodes in the MOEA/D citation network. Node size is proportional to number of citations, color indicates publication year. Edge width implies paper similarity based on co-citation and bibliographic coupling. Paper information can be found in \textsc{Appendix} C.}
    \label{fig:citation_core}
\end{figure}

In addition to community detection, another major task in bibliometric analysis is to trace a field's evolution through a \textit{main path} in the citation network. This path aims to highlight landmark studies and their interconnections~\cite{HummonD89,Batagelj03,LiuL12}. However, in domains such as MOEA/D, one path (or even citation trees~\cite{WuWE19}) may not capture the many-faceted relationships among papers. More influential subgraphs could reveal complex knowledge flows, yet become less interpretable as the number of edges grows.

To address these challenges, we propose a novel main path analysis composed of three steps.
\paragraph{Identifying essential works} We first apply the ranking method from~\cite{WangTZ13} to find a set of highly influential papers in the citation network. This method simultaneously considers:
$\blacktriangleright$ paper citations (including how influential the citing papers are), $\blacktriangleright$ author authority, and $\blacktriangleright$ venue prestige, using the knowledge graph in \pref{sec:data}. The algorithm also accounts for the network's dynamic nature to reduce bias toward earlier, established works. This produces more balanced rankings than Cite-Rank~\cite{WalkerXYM07} or P-Rank~\cite{YanDS11}.

\paragraph{Network trimming} Next, we extract a subgraph centered on these essential papers. We then allow papers that cite others to \lq inherit\rq\ their references. This removes redundant links to older works (e.g., MOEA/D) when they are already cited by the referenced paper. As a result, the network focuses on paths leading to new knowledge.

\paragraph{Edge weighting} To improve interpretability, we use co-citation and bibliographic coupling to weight edges in the trimmed subgraph. Co-citation arises when two papers are jointly cited by later work, suggesting an overlap in content. Bibliographic coupling occurs when two papers share references, indicating a shared background. We compute each pair's co-citations and the Jaccard similarity of their reference sets. Their normalized sum is the edge weight, where a larger weight denotes greater relevance and similarity.

\pref{fig:citation_core} shows the \lq backbone\rq\ of the MOEA/D citation network extracted by our method. It mainly consists of papers published within $10$ years of MOEA/D's inception that have established their influence. Overall, most works build on MOEA/D (2007)~\cite{ZhangL07} and NSGA-III (2014)~\cite{DebJ14,JainD14}. Instead of forming a neat linear progression, these studies interweave into a complex structure. One reason is that MOEA/D research often intersects the broader EMO domain, linking both areas. Notably, Hisao Ishibuchi's works (e.g.,~\cite{IshibuchiSMN17,IshibuchiSTN09}) frequently bridge different research streams. They rely on purpose-built experiments that deepen our understanding of EMO algorithms and spark new research directions.

\section{Collaboration Network Analysis}
\label{sec:collaboration}

In this section, we analyze the author collaboration network, another important component of the MOEA/D literature landscape. As in previous sections, we begin by examining various network metrics to understand the community's composition and evolution. We then highlight the most active researchers and explore how they collaborate.

\begin{figure}[t!]
    \centering
    \includegraphics[width=\linewidth]{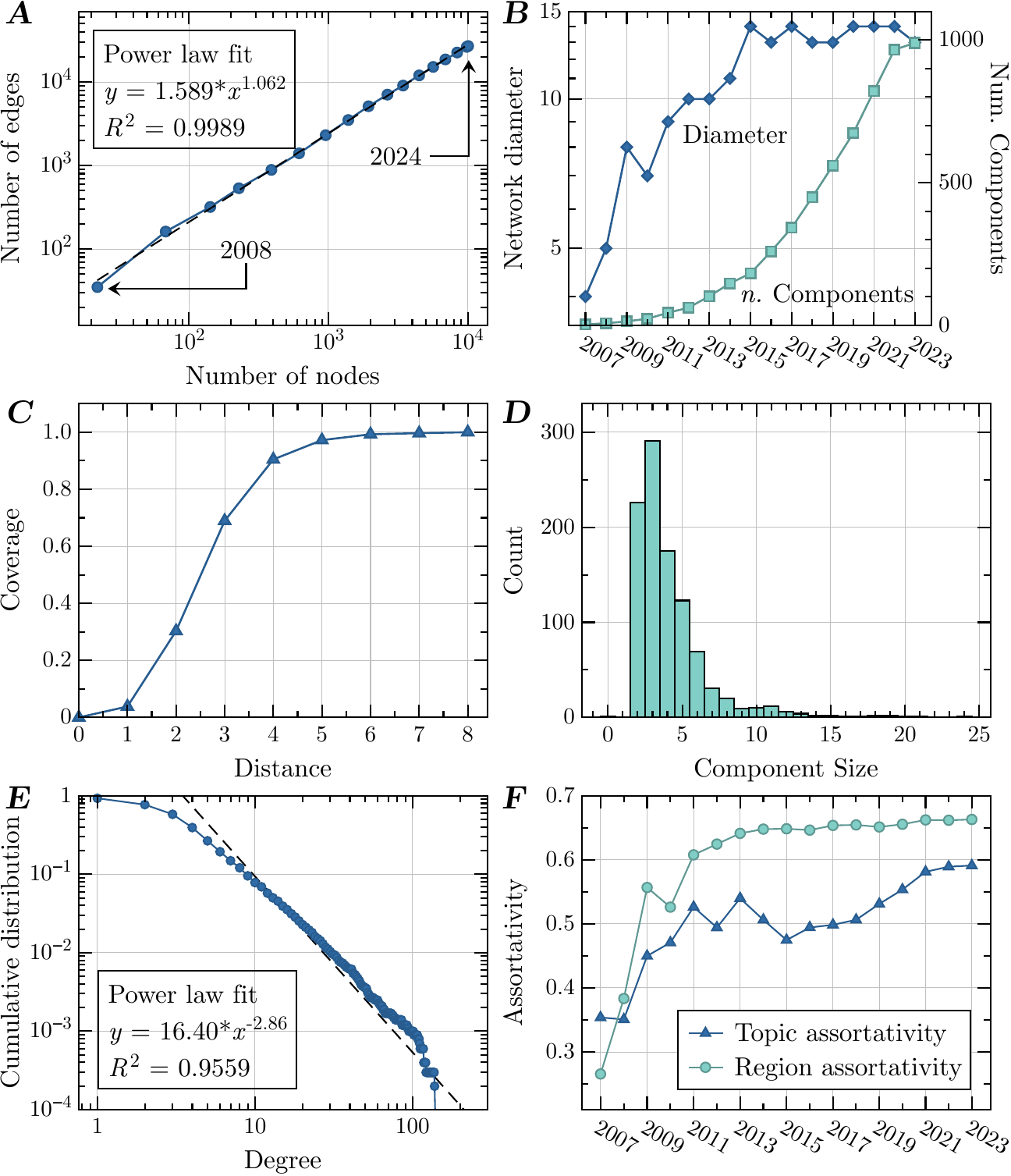}
    \caption{General patterns for MOEA/D author collaboration networks. \textbf{(A)} Network growth during 2008 and 2024. \textbf{(B)} Network diameter (largest connected component) and number of connected components each year. \textbf{(C)} Percentages of nodes that can be reached within different numbers of hubs from the highest-degree node in the largest connected component. \textbf{(D)} Distribution of the size of the connected components in the network, excluding the largest one. \textbf{(E)} Cumulative distribution of author connections within the network. \textbf{(F)} Assortativity coefficient for author nationality and primary research interest.}
    \label{fig:an_metrics}
\end{figure}

\pref{fig:an_metrics}(B) tracks the evolving number of connected components in the MOEA/D collaboration network. By the end of 2023, there are about $1,000$ isolated research groups, most with fewer than five members (\pref{fig:an_metrics}(D)) and focused on application-oriented research. The largest of these groups has $24$ researchers. In contrast, the network's main component contains more than $5,700$ authors. This disparity suggests a wide range of scales and research foci across the community.

\subsubsection{Network growth and small world effect}
As shown in~\pref{fig:an_metrics}(A), the MOEA/D community has expanded rapidly in recent years, following a power-law exponent of $\alpha=1.06$. This implies denser connections over time and leads to a small-world structure~\cite{Milgram67}. One indicator is the stable network diameter $d$, which has remained at about $14$ since 2015 despite continual growth (\pref{fig:an_metrics}(B)). In the largest component, $70\%$ of authors lie within just three hops of the highest-degree node, and nearly all are within five hops (\pref{fig:an_metrics}(C)). This compact connectivity suggests that the MOEA/D community is growing closer, making collaboration and knowledge sharing more efficient.

\subsubsection{Degree distribution}
\pref{fig:an_metrics}(E) shows that the author collaboration network has a highly skewed degree distribution, similar to many citation networks. Most authors connect to only a handful of colleagues, often reflecting a supervisor-student relationship or a small, tight-knit group. This abundance of low-degree nodes shapes the network's foundational structure and hints at a hierarchical, possibly insular collaboration model. It also highlights the role of highly connected individuals in linking these smaller groups.

\subsubsection{Mixing patterns}
Finally, we examined whether MOEA/D researchers prefer collaborating with others in the same geographic region or those studying similar topics. The assortativity coefficients~\cite{Newman10} in~\pref{fig:an_metrics}(F) show a rising trend in both geographical and topical homophily over time. This pattern suggests a shift toward more narrowly focused research circles and stable regional networks. While this may foster in-depth exploration within established research domains, it is expected to promote broader interdisciplinary and cross-regional partnerships, which have been shown to infuse fresh perspectives and foster innovation in a research community~\cite{LinFW23,PriceB69}.


\subsubsection{Collaboration patterns}
\begin{figure}[t!]
    \centering
    \includegraphics[width=.9\linewidth]{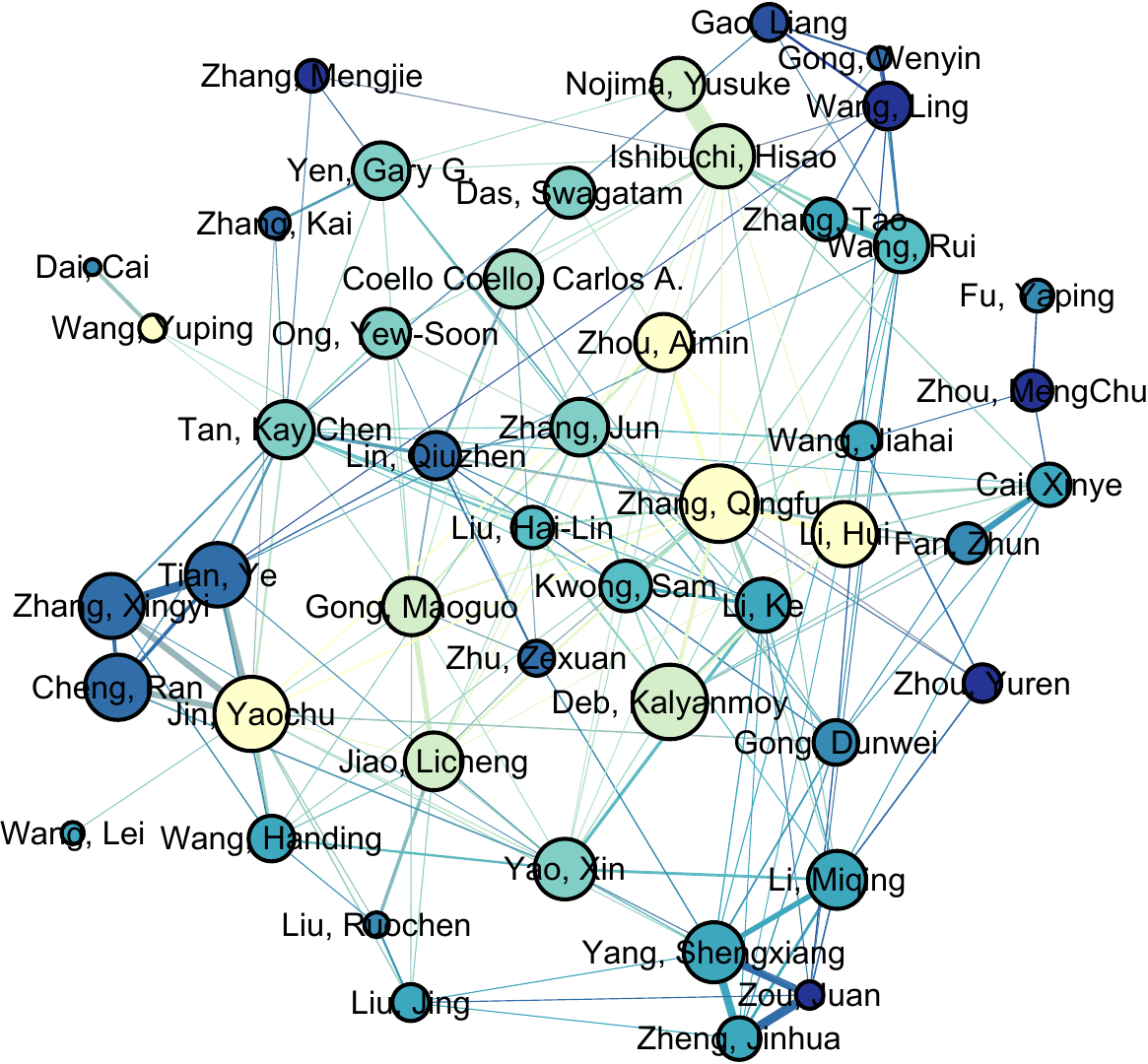}
    \caption{Local community of top-50 active authors in the MOEA/D community. Node size and color are proportional to the author's PageRank centrality and year of entry, respectively. Edge width implies collaboration strength. Full list of authors and be found in Appendix D.}
    \label{fig:author_comm}
\end{figure}

To gain insight into collaboration patterns in the MOEA/D community, we identified the $50$ most active researchers using PageRank centrality. This metric captures how actively a researcher collaborates with others. Note that it is not intend to assess publication quantity, quality, or reputation. As shown in~\pref{fig:author_comm}, the top active authors form a closely connected network. Notably, there are $26$ four-cliques and $14$ five-cliques. Authors who bridge these groups, such as Carlos A. Coello Coello and \textit{Prof.} Kay Chen Tan, tend to have high betweenness centrality. Central figures in the network, especially Qingfu Zhang, have strong collaborative ties with Hui Li, Aimin Zhou, Kalyanmoy Deb, Ke Li, and Sam Kwong. Within the whole network, the strongest tie is between Hisao Ishibuchi and Yusuke Nojima.


\section{Predicting Future Research}
\label{sec:prediction}

The previous sections have demonstrated the capability of \our\ for revealing hidden patterns within the extensive knowledge captured in the MOEA/D literature landscape. Just as our human scientists, we generate research ideas by reading and digesting existing literature. Therefore, we hypothesize that \our\ can help predict future MOEA/D research directions by identifying which keyword combinations may attract researchers. Specifically, we focus on a keyword co-occurrence network~\cite{RzhetskyFFE15}, where nodes are MOEA/D-related keywords (e.g., \lq hypervolume\rq, \lq diversity\rq), and edges form when a single paper contains both. We only consider keywords in titles or abstracts, as these typically capture core research ideas. For example, the title \lq Stable Matching-Based Selection in Evolutionary Multiobjective Optimization\rq~\cite{LiZKLW14} clearly signals a connection between stable matching algorithms and selection strategies in EMO.

\begin{figure}[t!]
    \centering
    \includegraphics[width=\linewidth]{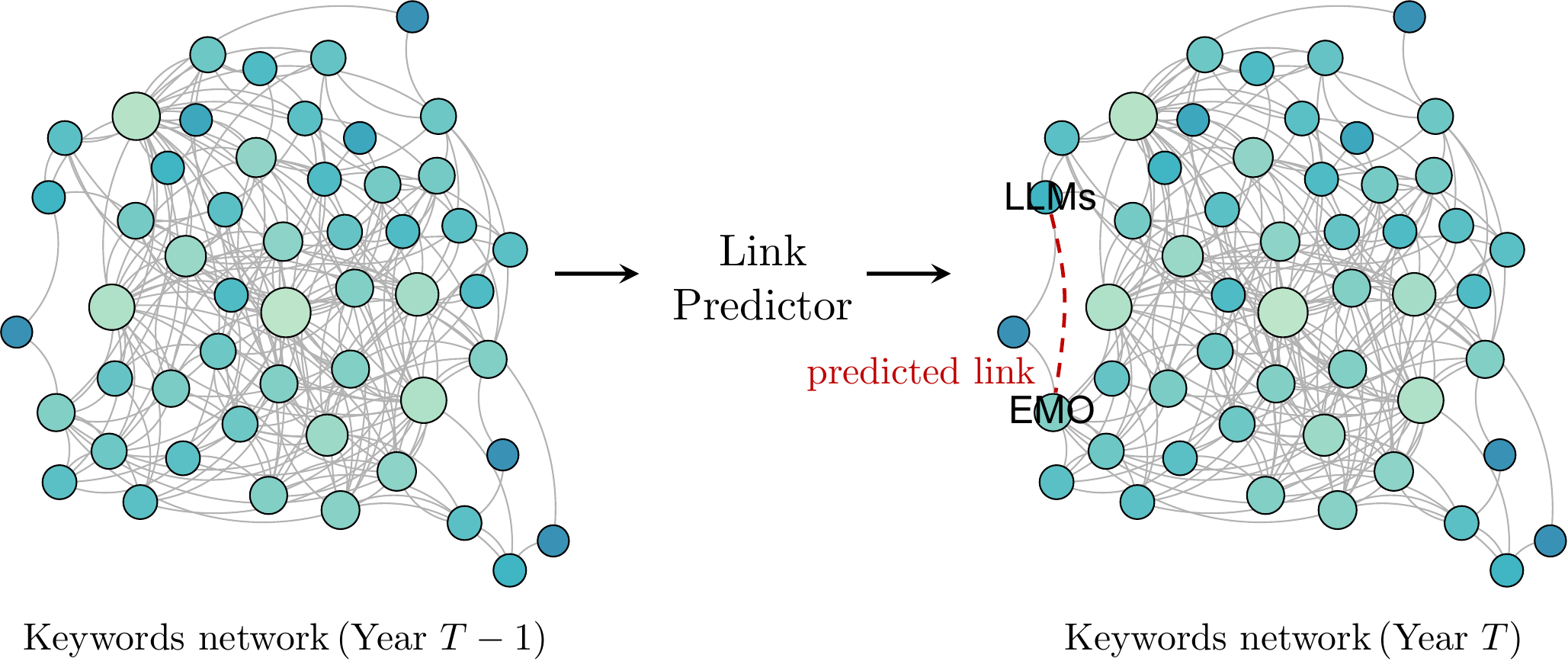}
    \caption{Illustration of the link prediction task in the keyword co-occurrence network. The model predicts which unconnected nodes, i.e., keywords that appeared in paper abstracts, are likely to be connected in the future based on current network typology.}
    \label{fig:prediction}
\end{figure}

As new research appears, this keyword network evolves, reflecting the field's history and scientists' collective focus. We treat future research prediction as a link-prediction task~\cite{MartinezBT17}. It aims to forecast which currently unconnected keywords will become linked. Our model combines engineered network features (see~\cite{Lu21}) and a gradient boosting tree classifier. Though relatively simple, this method outperformed alternatives in the \texttt{Science4Cast benchmark}~\cite{KrennBCC23}. We trained the model by extracting node pairs from an earlier snapshot (year $T-1$, for $T\in\{2010,2011,\dots,2024\}$) that either formed a new edge or remained unconnected in the next year. Each pair was labeled positive or negative based on whether they eventually connected. We then computed various structural features (e.g., degree centrality, shared neighbors, temporal trends) for these node pairs, and fed these into the gradient boosting classifier, verifying performance against the actual links that appeared.

After training, we applied the model to all unconnected pairs in 2023, ranking them by their probability of connecting in 2024. This setting allows us to verify the model's predictions with the actual data from $2024$. Among the top $100$ predicted links, we found that many of them were indeed studied in new works published last year. In particular, we observed many connections between \lq LLMs\rq\ and other EMO or MOEA/D keywords (e.g., \lq evolutionary computation\rq, \lq selection\rq, \lq Pareto dominance\rq). Our query found at least $85$ publications in WoS and more than $1,000$ \textit{arXiv} preprints exploring these ideas in 2024. For example, some studies discussed the potential of using LLMs to assist evolutionary algorithms~\cite{WuWFW24,vanSteinB24,LiuCCOT24}. On the other hand, LLM research has also been a new application field of EMO and EC. For instance, \cite{ZhangCWCHW25} studied the use of evolutionary algorithms to optimize multi-agent LLM systems, which has also been discussed in Li's latest survey~\cite{Li24}. In addition, \cite{QiQLGZB24} and \cite{ZhaoYPJYQ25} apply the idea of EMO for prompt optimization for LLMs. Furthermore, \cite{ShiCHLSD24} proposes a multi-objective approach for LLM alignment, and \cite{AkibaSTSHD25} applies evolutionary algorithms for LLM merging. 

The success of predicting research in 2024 suggests that analyzing literature landscapes could reveal early hints for future directions. Looking ahead, \our\ may thus serve as a powerful tool for accelerating research and innovation.

\section{Expert Interview}
\label{sec:interview}

To evaluate the practical utility of \our, we conducted interviews with $20$ domain experts actively involved in MOEA/D and EMO research. Among them were five senior experts, each with over a decade of research experience, and $15$ doctoral researchers currently engaged in MOEA/D-related projects. Each interview session lasted about one hour. We presented our analytical pipeline and key findings, then asked experts whether these insights exceeded their existing knowledge. An open discussion followed, allowing interactive system use and feedback on usability and potential enhancements.
\begin{enumerate}
    \item Most experts found the general analysis section highly valuable. In fact, $80\%$ explicitly praised its capacity to provide a clear overview of publication trends, key venues, researcher demographics, and citation intents. One senior expert remarked, \lq\lq\textit{This general overview efficiently summarizes crucial information that would otherwise require significant manual effort.}\rq\rq\ Another expert highlighted the clarity of the geographic visualization: \lq\lq\textit{The map clarifies global research contributions, especially the role of Chinese researchers.}\rq\rq

    \item Topic modeling analysis received the most positive recognition, with $95\%$ of experts calling it particularly insightful. Many participants were surprised by prevalent research topics they had not noticed before. As one senior expert said, \lq\lq\textit{The identification of emerging topics such as multi-modal optimization and preference-based multi-objective optimization offers valuable foresight into future research potentials.}\rq\rq\ A Ph.D. student also noted, \lq\lq\textit{Visualizing evolving trends helps direct my future research.}\rq\rq

    \item For citation network analysis, $75\%$ of participants appreciated the clear depiction of how influential studies interconnect over time. Senior experts valued the understanding of disruptiveness. One observed, \lq\lq\textit{Knowing which papers disrupt or consolidate existing knowledge allows for more informed judgments on significant contributions.}\rq\rq\ However, they suggested more contextual explanations for shifts in disruptiveness. One expert proposed, \lq\lq\textit{Future analysis could explore reasons behind these shifts. These will make results more actionable.}\rq\rq

    \item Collaboration network analysis impressed $90\%$ of experts, especially for revealing community dynamics and potential collaborators. One participant remarked, \lq\lq\textit{Analyzing collaboration patterns reveals critical hubs in the community, enabling targeted networking and partnerships.}\rq\rq\ Another expert pointed out its usefulness for newcomers: \lq\lq\textit{Understanding collaboration dynamics is essential for early-career researchers entering the field.}\rq\rq

    \item Finally, the predictive analytics section captured the interest of $80\%$ of participants. They praised its innovative approach to forecasting future research. Several experts noted its accuracy in identifying intersections between MOEA/D and LLMs. One participant said, \lq\lq\textit{It is impressive how accurately the prediction model identified emerging trends like EMO applications to LLM research, which aligns perfectly with current developments.}\rq\rq
\end{enumerate}

In summary, the interviewed experts unanimously recognized the value and novelty of our proposed \our. They highlighted its potential to improve literature exploration efficiency, coverage, and accuracy compared to SLRs. They also provided constructive suggestions for enhancing usability. For example, they recommended developing an interactive user interface to facilitate user engagement. Additionally, they encouraged contextualizing analysis results via LLMs to support broader adoption among researchers with diverse technical backgrounds.

\section{Concluding Remarks}
\label{sec:conclusion}

In this paper, we introduced \our, a data-driven landscape analysis tool for large-scale literature landscape exploration. We applied it to MOEA/D research, encompassing over $5,400$ papers, $10,000$ authors, $400$ venues, and $1,600$ institutions. Our findings provide a broad view of this field, placing insights from earlier expert surveys~\cite{Li24,XuXM20,TrivediSSG17,PinedaHDPSBM14,WangSLMGLM20,MaYLQZ20,Guo22a,XuXM19,MaYWJZ16} into a wider context. The prominence and diversity of MOEA/D research also mirror the rising importance of multi-objective optimization in real-world applications. Furthermore, our literature landscape analyses reveal patterns that align with recognized trends in scientometric and complex network research. Last but not least, we also demonstrated how the knowledge captured by \our\ can be leveraged to predict future research ideas.

There are some limitations within this study.
\begin{itemize}
    \item[\faBomb] We rely on papers citing MOEA/D~\cite{ZhangL07}, which may include works not directly tied to decomposition-based EMO. This approach might overestimate the observed research scope and topic prevalence. However, it remains justifiable within the broader EMO domain because our topic modeling also uncovers discussions related to NSGA. Further, due to the scale of the data and the ambiguity of paper contents, it is infeasible to define a subset of papers that are deemed to be \lq relevant\rq\ to MOEA/D, which is itself subjective.

    \item[\faBomb] The \texttt{BERTopic} model depends on hyperparameters, which may affect the results. By experimenting with different settings, we found major topics remain stable under similar resolutions. Smaller or less stable topics were refined with expert consultation.

    \item[\faBomb] Although we used advanced data mining methods, there is much potential for improvement. For instance, combining topic modeling with community detection could capture both content and structural insights. Predictive models might also better assess the long-term impact of papers or authors.
\end{itemize}

Despite these challenges, this paper provides a useful sketch of the vast decomposition-based EMO literature landscape. Our analysis can help guide future work within and beyond this field. We also hope it illustrates how data-driven literature landscape analysis may become increasingly important as academic literature continues to expand. Beyond MOEA/D, \our\ can be adapted to other EMO areas and beyond.


\bibliographystyle{IEEEtran}
\bibliography{IEEEabrv,moead}

\end{document}